\pdfoutput=1

\documentclass[11pt]{article}

\usepackage[]{acl}

\usepackage{times}
\usepackage{latexsym}
\usepackage{graphicx}
\usepackage[T1]{fontenc}

\usepackage{amsmath}
\usepackage{amsfonts}
\usepackage{bbold}

\usepackage{inconsolata}

\usepackage{algorithm}
\usepackage{algpseudocode}
\newcommand{\algorithmicbreak}{\textbf{break}}
\newcommand{\Break}{\State \algorithmicbreak}

\usepackage{booktabs}
\newcommand{\num}[2]{#1$_\textrm{#2}$}
\newcommand{\bnum}[2]{\textbf{#1}$_\textrm{#2}$}

\usepackage{soul}
\usepackage{xcolor}
\newcommand{\hlc}[2][yellow]{{%
    \colorlet{foo}{#1}%
    \sethlcolor{foo}\hl{#2}}%
}
\DeclareRobustCommand{\hlred}[1]{{\hlc[red!50]{- #1}}}
\DeclareRobustCommand{\hlblue}[1]{{\hlc[cyan!50]{+ #1}}}

\usepackage[utf8]{inputenc}

\usepackage{microtype}

\title{Low-Resource Task-Oriented Semantic Parsing \\ via Intrinsic Modeling}

\author{Shrey Desai \quad\quad Akshat Shrivastava \quad\quad Alexander Zotov \quad\quad Ahmed Aly \\
  Facebook \\
  \tt{\{shreyd, akshats, azotov, ahhegazy\}@fb.com}}

\begin{document}
\maketitle

\begin{abstract}
Task-oriented semantic parsing models typically have high resource requirements: to support new ontologies (i.e., intents and slots), practitioners crowdsource thousands of samples for supervised fine-tuning. Partly, this is due to the structure of de facto copy-generate parsers; these models treat ontology labels as discrete entities, relying on parallel data to \textit{extrinsically} derive their meaning. In our work, we instead exploit what we \textit{intrinsically} know about ontology labels; for example, the fact that \texttt{SL:TIME\_ZONE} has the categorical type ``slot'' and language-based span ``time zone''. Using this motivation, we build our approach with offline and online stages. During preprocessing, for each ontology label, we extract its intrinsic properties into a component, and insert each component into an inventory as a cache of sorts. During training, we fine-tune a seq2seq, pre-trained transformer to map utterances and inventories to frames, parse trees comprised of utterance and ontology tokens. Our formulation encourages the model to consider ontology labels as a union of its intrinsic properties, therefore substantially bootstrapping learning in low-resource settings. Experiments show our model is highly sample efficient: using a low-resource benchmark derived from TOPv2 \cite{chen-2020-topv2}, our inventory parser outperforms a copy-generate parser by +15 EM absolute (44\% relative) when fine-tuning on 10 samples from an unseen domain.
\end{abstract}

\section{Introduction}
\label{sec:introduction}

Task-oriented conversational assistants face an increasing demand to support a wide range of domains (e.g., reminders, messaging, weather) as a result of their emerging popularity \cite{chen-2020-topv2,ghoshal-2020-loras}. For practitioners, enabling these capabilities first requires training semantic parsers which map utterances to frames executable by assistants \cite{gupta-2018-top,einolghozati-2018-improving,pasupat-2019-span,aghajanyan-2020-decoupled,li-2020-mtop,chen-2020-topv2,ghoshal-2020-loras}. However, current methodology typically requires crowdsourcing thousands of samples for each domain, which can be both time-consuming and cost-ineffective at scale \cite{wang-2015-overnight,jia-2016-recombination,herzig-2019-datacollection,chen-2020-topv2}. One step towards reducing these data requirements is improving the sample efficiency of current, de facto copy-generate parsers. However, even when leveraging pre-trained transformers, these models are often ill-equipped to handle low-resource settings, as they fundamentally lack inductive bias and cross-domain reusability.

In this work, we explore a task-oriented semantic parsing model which leverages the intrinsic properties of an ontology to improve generalization. To illustrate, consider the ontology label \texttt{SL:TIME\_ZONE} : a slot representing the time zone in a user's query. Copy-generate models typically treat this label as a discrete entity, relying on parallel data to \textit{extrinsically} learn its semantics. In contrast, our model exploits what we \textit{intrinsically} know about this label, such as its categorical type (e.g., ``slot'') and language-based span (e.g., ``time zone''). Guided by this principle, we extract the properties of each label in a domain's ontology, building a component with these properties and inserting each component into an inventory. By processing this domain-specific inventory through strong language models, we effectively synthesize an inductive bias useful in low-resource settings.

\begin{figure*}
    \centering
    \includegraphics[scale=0.26]{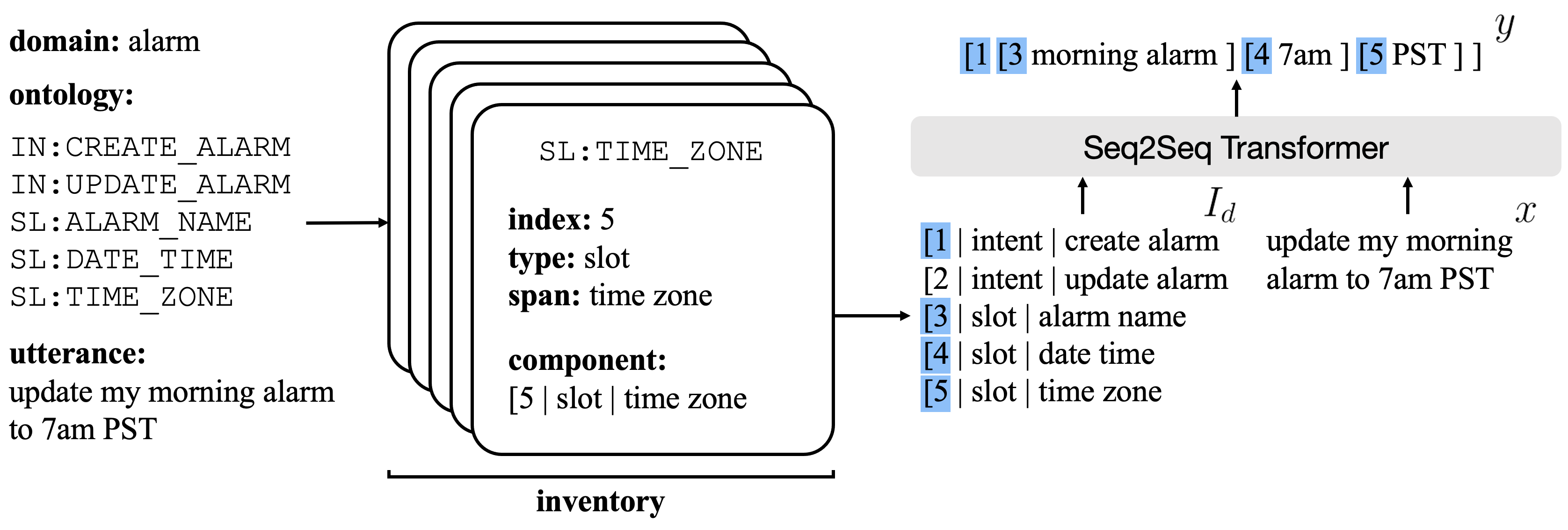}
    \caption{\textbf{Illustration of our low-resource, task-oriented semantic parser.} Let $x$ represent an utterance and $d$ the utterance's domain, which consists of an ontology (list of intents and slots) $\ell^d_1, \cdots, \ell^d_m$. We create an inventory $\mathbf{I}_d$, where each component contains intrinsic properties (index, type, span) derived from its respective label. Then, we fine-tune a seq2seq transformer to input a linearized inventory $I_d$ and utterance $x$ and output a frame $y$. Here, frames are composed of utterance and ontology tokens; ontology tokens, specifically, are referenced naturally via encoder self-attention rather than an augmented decoder vocabulary, like with copy-generate mechanisms.}
    \label{fig:model}
\end{figure*}

Concretely, we build our model on top of seq2seq, pre-trained transformer, namely BART \cite{lewis-2020-bart}, and fine-tune it to map utterances to frames, as depicted in Figure~\ref{fig:model}. Our model operates in two stages: (1) the encoder inputs a domain-specific utterance and inventory and (2) the decoder outputs a frame composed of utterance and ontology tokens. Here, instead of performing vocabulary augmentation, the standard way of ``generating'' ontology tokens in copy-generate parsers, we treat ontology tokens as pointers to inventory components. This is particularly useful in low-resource settings; our model is encouraged to represent ontology tokens as a union of its intrinsic properties, and as a result, does not require many labeled examples to achieve strong performance.

We view sample efficiency as one of the advantages of our approach. As such, we develop a comprehensive low-resource benchmark derived from TOPv2 \cite{chen-2020-topv2}, a task-oriented semantic parsing dataset spanning 8 domains. Using a leave-one-out setup on each domain, models are fine-tuned on a high-resource, \textit{source} dataset (other domains; ~100K+ samples), then fine-tuned and evaluated on a low-resource, \textit{target} dataset (this domain; 10-250 samples). We also randomly sample target subsets to make the transfer task more difficult. In aggregate, our benchmark provides 32 experiments, each varying in domain and number of target samples. 

Both coarse-grained and fine-grained experiments show our approach outperforms baselines by a wide margin. Overall, when averaging across all domains, our inventory model outperforms a copy-generate model by +15 EM absolute (44\% relative) in the most challenging setting, where only 10 target samples are provided. Notably, our base inventory model (139M parameters) also outperforms a large copy-generate model (406M parameters) in most settings, suggesting usability in resource-constrained environments. We also show systematic improvements on a per-domain basis, even for challenging domains with high compositionality and ontology size, such as reminders and navigation. Finally, through error analysis, we show our model's predicted frames are largely precise and linguistically consistent; even when inaccurate, our frames not require substantial modifications to achieve gold quality.

\section{Background and Motivation}
\label{sec:background-and-motivation}

Task-oriented semantic parsers typically cast parsing as transduction, utilizing seq2seq transformers to map utterances to frames comprised of intents and slots \cite{aghajanyan-2020-decoupled,li-2020-mtop,chen-2020-topv2,ghoshal-2020-loras}. Because frames are a composition of utterance and ontology tokens, these models are often equipped with copy-generate mechanisms; at each timestep, the decoder either \textit{copies} from the utterance or \textit{generates} from the ontology \cite{see-2017-copy-gen,aghajanyan-2020-decoupled,chen-2020-topv2,ghoshal-2020-loras}. These parsers are empirically effective in high-resource settings, achieving state-of-the-art performance on numerous benchmarks \cite{aghajanyan-2020-decoupled}, but typically lack inductive bias in low-resource settings.

To illustrate, consider a hypothetical domain adaptation scenario where a copy-generate parser adapts to the weather domain. In standard methodology, a practitioner augments the decoder's vocabulary with weather ontology labels, then fine-tunes the parser on weather samples. This subsequently trains the copy-generate mechanism to \textit{generate} these labels as deemed appropriate. But, the efficacy of this process scales with the amount of training data as these ontology labels (though, more specifically, their embeddings) iteratively derive \textbf{extrinsic} meaning. Put another way, before training, there exists no correspondence between an ontology label (e.g., \texttt{SL:LOCATION}) and an utterance span (e.g., ``Menlo Park''). Such an alignment is only established once the model has seen enough parallel data where the two co-occur.

In contrast, we focus on reducing data requirements by exploiting the \textbf{intrinsic} properties of ontology labels. These labels typically have several core elements, such as their label types and spans. For example, by teasing apart \texttt{SL:LOCATION}, we can see it is composed of the type ``slot'' and span ``location''. These properties, when pieced together and encoded by strong language models, provide an accurate representation of what the ontology label \textit{is}. While these properties are learned empirically with sufficient training data, as we see with the copy-generate parser, our goal is to train high-quality parsers with as little data as possible by explicitly supplying this information. Therefore, a central question of our work is whether we can build a parser which leverages the intrinsic nature of an ontology space while retaining the flexibility of seq2seq modeling; we detail our approach in the next section.

\section{Semantic Parsing via Label Inventories}
\label{sec:semantic-parsing-via-label-inventories}

\begin{table}[t]
\centering
\setlength{\tabcolsep}{3.5pt}
\begin{tabular}{lrrr}
\toprule
& \multicolumn{3}{c}{Components} \\
\cmidrule(lr){2-4}
Ontology Label & Index & Type & Span \\
\midrule
$\texttt{IN:CREATE\_ALARM}$ & 1 & intent & create alarm \\
$\texttt{IN:UPDATE\_ALARM}$ & 2 & intent & update alarm \\
$\texttt{SL:ALARM\_NAME}$ & 3 & slot & alarm name \\
$\texttt{SL:DATE\_TIME}$ & 4 & slot & date time \\
$\texttt{SL:TIME\_ZONE}$ & 5 & slot & time zone \\
\bottomrule
\end{tabular}
\caption{Example inventory $\mathbf{I}_\textrm{alarm}$ (right-half) with components $\mathbf{c}_{1:5}$, each corresponding to ontology labels (left-half) from the alarm domain.}
\label{tab:inventory_alarm}
\end{table}

Illustrated in Figure~\ref{fig:model}, we develop a seq2seq parser which uses \textbf{inventories}---that is, tables enumerating the intrinsic properties of ontology labels---to map utterances to frames. Inventories are domain-specific, and each \textbf{component} carries the intrinsic properties of a single label in the domain's ontology. On the source-side, a pre-trained encoder consumes both an utterance and inventory (corresponding to the utterance's domain). Then, on the target-side, a pre-trained decoder mimics a copy-generate mechanism by either selecting from the utterance or ontology. Instead of selecting ontology labels from an augmented vocabulary, as in copy-generate methods, our decoder naturally references these labels in the source-side inventory through self-attention. The sequence ultimately decoded during generation represents the frame.

As alluded to earlier, we focus on two intrinsic properties of ontology labels: \textbf{types} and \textbf{spans}. The type is particularly useful for enforcing syntactic structure; for example, the rule ``slots cannot be nested in other slots'' \cite{gupta-2018-top} is challenging to meet unless a model can delineate between intents and slots. Furthermore, the span is effectively a natural language description, which provides a general overview of what the label aligns to. Though inventory components can incorporate other intrinsic properties, types and spans do not require manual curation and can be automatically sourced from existing annotations.

Despite the reformulation of the semantic parsing task with inventories, our approach inherits the flexibility and simplicity of copy-generate models \cite{aghajanyan-2020-decoupled,li-2020-mtop,chen-2020-topv2,ghoshal-2020-loras}; we also treat parsing as transduction, leverage pre-trained modules, and fine-tune with log loss. However, a key difference is our parser is entirely text-to-text and does not require extra parameters, which we show promotes reusability in low-resource settings. In the following sub-sections, we elaborate on our approach in more detail and comment on several design decisions.

\subsection{Inventories}
\label{sec:inventories}

Task-oriented semantic parsing datasets typically have samples of the form $(d, x, y)$ \cite{gupta-2018-top,li-2020-mtop,chen-2020-topv2}; that is, a domain $d$, utterance $x$, and frame $y$. There exist many domains $d_1, \cdots, d_n$, where each domain defines an ontology $\ell^{d}_1, \cdots, \ell^{d}_m$, or list of intents and slots. For a given domain $d$, we define its inventory as a table $\mathbf{I}_d = [\mathbf{c}^{d}_1, \cdots, \mathbf{c}^{d}_m]$ where each component $\mathbf{c}^{d}_i = (i, t, s)$ is a tuple storing the intrinsic properties of a corresponding label $\ell^{d}_i$. 

Specifically, these components consist of: (1) an \textbf{index} $i \in \mathbb{Z}^{\ge}$ representing the label's position in the (sorted) ontology; (2) a \textbf{type} $t \in \{\textrm{intent}, \textrm{slot}\}$ denoting whether the label is an intent or slot; and (3) a \textbf{span} $s \in V^{*}$ representing an ontology description, formally represented as a string from a vocabulary $V$. The index is a unique referent to each component and is largely used as an optimization trick during generation; we elaborate on this in the next section. In Table~\ref{tab:inventory_alarm}, we show an example inventory for the alarm domain.

\subsection{Seq2Seq Model}
\label{sec:seq2seq-model}

Our model is built on top of a pre-trained, seq2seq transformer architecture \cite{vaswani-2017-transformer} with vocabulary $V$.

\paragraph{Encoder.} The \textbf{input} to the model is a concatenation of an utterance $x \in V$ and its domain $d$'s inventory $I_d \in V$.

Following recent work in tabular understanding \cite{yin-2020-tabert}, we encode our tabular inventory $\mathbf{I}_d$ as a linearized string $I_d$. As shown in Figure~\ref{fig:model}, for each component, the index is preceded by \texttt{[} and the remaining elements are demarcated by \texttt{|}. Because our tabular structure is not significantly complex, we elect not to use explicit row and column segment embeddings.

\paragraph{Decoder.} The \textbf{output} from the model is a frame $y \in V$, where at timestep $t$, the decoder either selects an utterance token ($y_t \in x$) or ontology token ($y_t \in \mathbb{Z}^{\ge}$).

Here, we use each component's index in place of typical ontology tokens. Similar to when a copy-generate parser \textit{generates} a token from an ontology, our inventory parser \textit{generates} an index corresponding to an entry. A key advantage is these indices, numerical values by nature, are already present in most transformer vocabularies and therefore do not require special augmentation. We primarily use this format to minimize the target sequence length; instead of requiring the decoder to generate a label's intrinsic properties as a means of ``selecting'' it, which typically requires several decoding steps, we use the label's index as a proxy. Implicitly, this manifests in a pooling effect during training, where the index acts as a snapshot over the corresponding component.

Furthermore, because our gold frames do not originally come with index pointers, we modify these frames to ensure compatibility with our approach. Implementation-wise, we maintain a dictionary of indices to ontology labels, which ensures this mapping is injective.

\paragraph{Optimization.} Finally, we fine-tune our seq2seq model by minimizing the log loss of the gold frame token at each timestep, conditioning on the utterance, inventory, and previous timesteps:
\begin{equation*}
    \mathcal{L}(\theta) = -\sum_{(d, x, y)} \sum_t \log P(y_t | x, I_d, y_{<t};\theta)
\end{equation*}

\section{Low-Resource Semantic Parsing Benchmark}
\label{sec:low-resource-semantic-parsing-benchmark}

\begin{table}[t]
\centering
\setlength{\tabcolsep}{4pt}
\begin{tabular}{lrrrrr}
\toprule
Domains & \multicolumn{1}{c}{Source} & \multicolumn{4}{c}{Target (SPIS)} \\
 \cmidrule(lr){3-6}
 &  & 1 & 2 & 5 & 10 \\
\midrule
Alarm & 104,167 & 13 & 25 & 56 & 107 \\
Event & 115,427 & 22 & 33 & 81 & 139 \\
Messaging & 114,579 & 23 & 44 & 89 & 158 \\
Music & 113,034 & 19 & 41 & 92 & 187 \\
Navigation & 103,599 & 33 & 63 & 141 & 273 \\
Reminder & 106,757 & 34 & 59 & 130 & 226 \\
Timer & 113,073 & 13 & 27 & 62 & 125 \\
Weather & 101,543 & 10 & 22 & 47 & 84 \\
\bottomrule
\end{tabular}
\caption{\textbf{TOPv2-DA benchmark training splits.} Models are initially fine-tuned on a source dataset, then fine-tuned on an SPIS subset from a target dataset.}
\label{tab:benchmark}
\end{table}

In this section, we describe our low-resource benchmark used to assess the sample efficiency of our model. The benchmark is derived from TOPv2 \cite{chen-2020-topv2}, a task-oriented semantic parsing dataset covering 8 domains: alarm, event, messaging, music, navigation, reminder, timer, and weather. TOPv2 samples have a combination of both linear and nested frames, uniquely reflecting the data distribution our parsers are likely to encounter in practice.

\paragraph{TOPv2-DA Benchmark.} To build our benchmark, nicknamed TOPv2-DA\footnote{TOPv2 Domain Adaptation (DA)}, we adopt a paradigm of source and target dataset fine-tuning, where a model is initially fine-tuned on a \textbf{high-resource, source dataset} (consisting of multiple domains), and is then fine-tuned on a \textbf{low-resource, target dataset} (consisting of one domain). This process describes one such transfer \textbf{scenario}; within this scenario, we can assess a model's few-shot capabilities incrementally by fine-tuning it on multiple \textbf{subsets}, each randomly sampled from the target dataset.

Table~\ref{tab:benchmark} provides a quantitative overview of our benchmark. We typically have 100K+ samples for source fine-tuning, but only about 10-250 samples for target fine-tuning, depending on the subset used. In aggregate, our benchmark provides 32 experiments (8 scenarios $\times$ 4 subsets), offering a rigorous evaluation of sample efficiency. 

\paragraph{Creating Experiments.} We use a leave-one-out algorithm to create source and target datasets. Given domains $\{d_1, \cdots, d_n\}$, we create $n$ scenarios where the $i$th scenario uses domains $\{d_j : d_i \ne d_j\}$ as the source dataset and domain $\{d_i\}$ as the target dataset. For each target dataset, we also create $m$ subsets using a random sampling algorithm, each with an increasing number of samples.

\begin{algorithm}
\caption{SPIS algorithm \cite{chen-2020-topv2}}
\begin{algorithmic}[1]
\State Input: dataset $D = \{(d^{(i)}, x^{(i)}, y^{(i)})\}^n_{i=1}$, subset cardinality $k$
\Procedure{SPIS}{$D$, $k$}
\State Shuffle $D$ using a fixed seed
\State $S \gets$ subset of dataset samples
\State $C \gets$ counter of ontology tokens
\For{$(d^{(i)}, x^{(i)}, y^{(i)}) \in D$}
    \For{ontology token $t \in y^{(i)}$}
        \If{$C\lbrack t\rbrack < k$}
            \State $S \gets S + (d^{(i)}, x^{(i)}, y^{(i)})$
            \State Store $y_i$'s ontology token
            \State \hspace{1em} counts in $C$
            \Break
        \EndIf
    \EndFor
\EndFor
\EndProcedure
\end{algorithmic}
\end{algorithm}

For our random sampling algorithm, we use \textbf{samples per intent slot (SPIS)}, shown abovex, which ensures at least $k$ ontology labels (i.e., intents and slots) appear in the resulting subset \cite{chen-2020-topv2}. Unlike a traditional algorithm which selects $k$ samples exactly, SPIS guarantees coverage over the entire ontology, but as a result, the number of samples per subset is typically much greater than $k$ \cite{chen-2020-topv2}. Therefore, we use conservative values of $k$; for each scenario, we sample target subsets of 1, 2, 5, and 10 SPIS. Our most extreme setting of 1 SPIS is still 10$\times$ smaller than the equivalent setting in prior work \cite{chen-2020-topv2,ghoshal-2020-loras}.

\section{Experimental Setup}
\label{sec:experimental-setup}

We seek to answer three questions in our experiments: (1) How sample efficient is our model when benchmarked on TOPv2-DA? (2) Does our model perform well on average or does it selectively work on particular domains? (3) How do the intrinsic components of an inventory component (e.g., types and spans) contribute to performance?

\paragraph{Systems for Comparison.} We chiefly experiment with CopyGen and Inventory, a classical copy-generate parser and our proposed inventory parser, as discussed in Sections \ref{sec:background-and-motivation} and \ref{sec:semantic-parsing-via-label-inventories}, respectively. Though both models are built on top of off-the-shelf, seq2seq transformers, the copy-generate parser requires special tweaking; to prepare its ``generate'' component, we augment the decoder vocabulary with dataset-specific ontology tokens and initialize their embeddings randomly, as is standard practice \cite{aghajanyan-2020-decoupled,chen-2020-topv2,li-2020-mtop}. In addition, both models are initialized with pre-trained weights. We use BART \cite{lewis-2020-bart}, a seq2seq transformer pre-trained with a denoising objective for generation tasks. Specifically, we we use the BART$_\textrm{BASE}$ (139M parameters; 12L, 768H, 16A) and BART$_\textrm{LARGE}$ (406M parameters; 24L, 1024H, 16A) checkpoints.

We benchmark the sample efficiency of these models on TOPv2-DA. Following the methodology outlined in Section~\ref{sec:low-resource-semantic-parsing-benchmark}, for each scenario and subset experiment, each model undergoes two rounds of fine-tuning: it is initially fine-tuned on a high-resource, source dataset, then fine-tuned again on a low-resource, target dataset using the splits in Table~\ref{tab:benchmark}. The resulting model is then evaluated on the target domain's TOPv2 test set; note that this set is \textit{not} subsampled for accurate evaluation. We report the exact match (EM) between the predicted and gold frame. To account for variance, we average EM across three runs, each with a different seed.

\paragraph{Hyperparameters.} We use BART checkpoints from \texttt{fairseq} \cite{ott-2019-fairseq} and elect to use most hyperparameters out-of-the-box. However, during initial experimentation, we find the batch size, learning rate, and dropout settings to heavily impact performance, especially for target fine-tuning. For source fine-tuning, our models use a batch size of 16, dropout in [0, 0.5], and learning rate in [1e-5, 3e-5]. Each model is fine-tuned on a single 32GB GPU given the size of the source datasets. For target fine-tuning, our models use a batch size in [1, 2, 4, 8], dropout in [0, 0.5], and learning rate in [1e-6, 3e-5]. Each model is fine-tuned on a single 16GB GPU. Finally, across both source and target fine-tuning, we optimize models with Adam \cite{kingma-2014-adam}.

\begin{table}
\centering
\setlength{\tabcolsep}{4pt}
\begin{tabular}{lrrrr}
\toprule
& \multicolumn{4}{c}{SPIS} \\
\cmidrule(lr){2-5}
 & 1 & 2 & 5 & 10 \\
\midrule
CopyGen$_\textrm{BASE}$ & 27.93 & 39.12 & 46.23 & 52.51 \\
CopyGen$_\textrm{LARGE}$ & 35.51 & 44.40 & 51.32 & 56.09 \\
Inventory$_\textrm{BASE}$ & 38.93 & 48.98 & 57.51 & 63.19 \\
Inventory$_\textrm{LARGE}$ & \textbf{51.34} & \textbf{57.63} & \textbf{63.06} & \textbf{68.76} \\
\bottomrule
\end{tabular}
\caption{\textbf{Coarse-grained results on TOPv2-DA.} Each model's EMs are averaged across 8 domains. Both Inventory$_\textrm{BASE}$ and Inventory$_\textrm{LARGE}$ outperform CopyGen in 1, 2, 5, and 10 SPIS settings.}
\label{tab:topv2da-average-results}
\end{table}

\begin{table*}
\centering
\setlength{\tabcolsep}{4pt}
\begin{tabular}{lrrrrrrrr}
\toprule
 & \multicolumn{4}{c}{Base Model (SPIS)} & \multicolumn{4}{c}{Large Model (SPIS)} \\
 \cmidrule(lr){2-5} \cmidrule(lr){6-9}
 & 1 & 2 & 5 & 10 & 1 & 2 & 5 & 10 \\
\midrule
\multicolumn{9}{l}{Domain: Alarm} \\
\midrule
CopyGen & \num{20.41}{1.16} & \num{38.90}{4.92} & \num{45.50}{4.00} & \num{52.01}{2.68} & \num{36.91}{3.10} & \num{43.70}{4.73} & \num{45.73}{1.42} & \num{53.89}{0.58} \\
Inventory & \bnum{62.13}{0.42} & \bnum{65.26}{0.94} & \bnum{71.81}{1.36} & \bnum{75.27}{2.16} & \bnum{67.25}{0.86} & \bnum{72.11}{0.96} & \bnum{71.82}{2.83} & \bnum{78.15}{2.04} \\
\midrule
\multicolumn{9}{l}{Domain: Event} \\
\midrule
CopyGen & \num{31.85}{0.22} & \num{38.85}{0.84} & \num{38.31}{0.71} & \num{41.78}{2.00} & \num{32.37}{0.72} & \num{34.59}{0.62} & \num{38.48}{0.21} & \num{43.93}{0.41} \\
Inventory & \bnum{46.57}{1.64} & \bnum{54.31}{0.53} & \bnum{58.87}{5.03} & \bnum{68.42}{2.06} & \bnum{64.77}{1.41} & \bnum{55.84}{4.20} & \bnum{67.70}{0.38} & \bnum{71.21}{1.08} \\
\midrule
\multicolumn{9}{l}{Domain: Messaging} \\
\midrule
CopyGen & \num{38.12}{0.61} & \num{49.79}{2.26} & \num{52.79}{2.56} & \num{58.90}{2.29} & \num{46.57}{4.70} & \num{58.42}{2.80} & \num{56.54}{7.94} & \num{63.10}{1.90} \\
Inventory & \bnum{46.54}{4.27} & \bnum{57.43}{2.48} & \bnum{63.72}{5.82} & \bnum{70.14}{3.98} & \bnum{60.36}{3.38} & \bnum{66.68}{3.45} & \bnum{74.69}{2.17} & \bnum{78.04}{2.46} \\
\midrule
\multicolumn{9}{l}{Domain: Music} \\
\midrule
CopyGen & \bnum{25.58}{0.19} & \num{33.28}{2.15} & \num{48.75}{2.07} & \bnum{55.16}{3.34} & \num{23.84}{17.42} & \num{36.84}{3.78} & \num{56.17}{0.55} & \num{59.18}{0.35} \\
Inventory & \num{23.00}{0.65} & \bnum{39.65}{4.48} & \bnum{53.59}{0.45} & \num{52.18}{2.49} & \bnum{38.68}{1.14} & \bnum{52.75}{1.71} & \bnum{58.23}{1.54} & \bnum{59.73}{1.73} \\
\midrule
\multicolumn{9}{l}{Domain: Navigation} \\
\midrule
CopyGen & \num{19.96}{0.84} & \bnum{30.11}{3.88} & \bnum{43.38}{1.24} & \num{45.26}{4.59} & \num{24.31}{0.97} & \num{36.28}{2.29} & \num{48.71}{1.75} & \num{56.14}{0.53} \\
Inventory & \bnum{21.16}{6.59} & \num{29.08}{0.47} & \num{42.59}{3.48} & \bnum{53.97}{0.56} & \bnum{28.74}{2.11} & \bnum{47.47}{2.42} & \bnum{49.98}{2.13} & \bnum{64.08}{0.88} \\
\midrule
\multicolumn{9}{l}{Domain: Reminder} \\
\midrule
CopyGen & \num{23.66}{3.18} & \num{23.30}{2.24} & \num{36.37}{2.64} & \num{41.66}{1.46} & \num{31.74}{2.53} & \num{31.82}{2.34} & \num{41.57}{1.73} & \num{42.62}{2.09} \\
Inventory & \bnum{28.58}{2.28} & \bnum{38.21}{2.81} & \bnum{48.88}{1.18} & \bnum{52.04}{5.76} & \bnum{40.72}{2.22} & \bnum{41.95}{2.25} & \bnum{53.57}{0.73} & \bnum{58.24}{2.75} \\
\midrule
\multicolumn{9}{l}{Domain: Timer} \\
\midrule
CopyGen & \num{16.62}{1.50} & \num{40.80}{21.47} & \num{54.79}{2.55} & \num{63.26}{0.85} & \num{32.64}{0.33} & \num{59.94}{1.73} & \num{59.80}{3.10} & \num{66.27}{2.54} \\
Inventory & \bnum{28.92}{1.95} & \bnum{53.58}{3.44} & \bnum{55.54}{3.90} & \bnum{66.82}{0.15} & \bnum{48.45}{4.44} & \bnum{61.70}{3.04} & \bnum{63.74}{1.09} & \bnum{68.44}{2.09} \\
\midrule
\multicolumn{9}{l}{Domain: Weather} \\
\midrule
CopyGen & \num{47.24}{11.10} & \bnum{57.97}{2.35} & \num{49.94}{11.30} & \num{62.07}{2.17} & \num{53.08}{1.31} & \num{53.60}{0.43} & \num{63.56}{3.41} & \num{63.58}{1.60} \\
Inventory & \bnum{54.53}{1.94} & \num{54.31}{2.87} & \bnum{65.09}{1.71} & \bnum{66.66}{3.40} & \bnum{61.77}{1.71} & \bnum{62.52}{1.85} & \bnum{64.73}{2.20} & \bnum{72.14}{1.76} \\
\bottomrule
\end{tabular}
\caption{\textbf{Fine-grained results on TOPv2-DA.} For each domain, base model EM is shown in the left-half and large model EM is shown in the right-half. Subscripts show standard deviation across three runs. Even on a per-domain basis, both Inventory$_\textrm{BASE}$ and Inventory$_\textrm{LARGE}$ outperform CopyGen in most 1, 2, 5, and 10 SPIS settings.}
\label{tab:topv2da-domain-results}
\end{table*}

\section{Results and Discussion}
\label{sec:results-and-discussion}

\subsection{TOPv2-DA Experiments}
\label{sec:topv2-da-experiments}

\begin{figure}[t]
    \centering
    \includegraphics[scale=0.5]{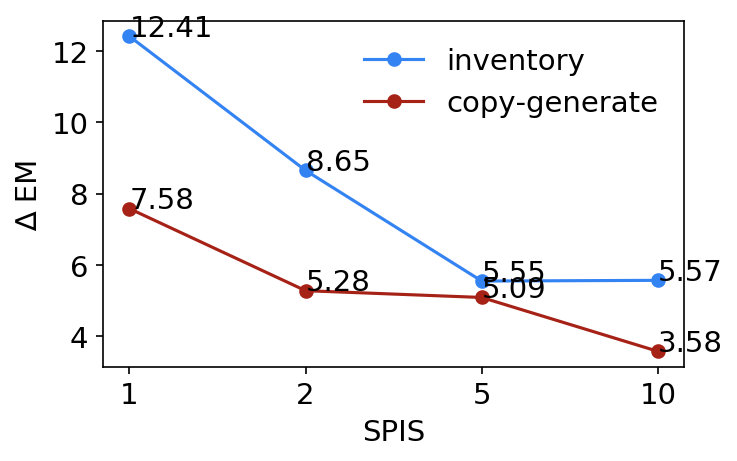}
    \caption{$\Delta$ EM between the base and large variants of Inventory and CopyGen on TOPv2-DA. Notably, Inventory makes the best use of large representations, with $\Delta$ = 12.41 EM at 1 SPIS.}
    \label{fig:parser_gap}
\end{figure}

\begin{figure}[t]
    \centering
    \includegraphics[scale=0.6]{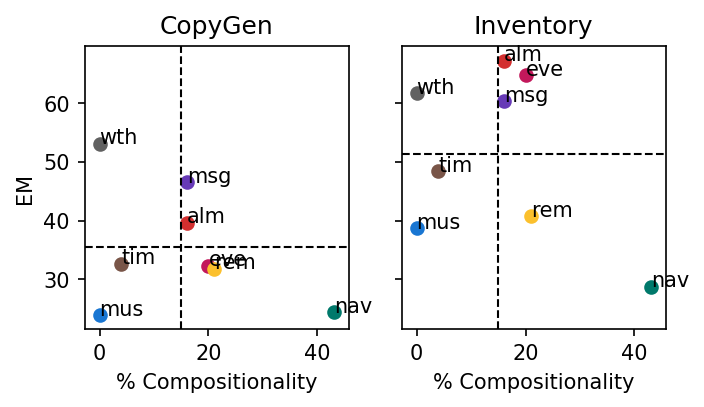}
    \includegraphics[scale=0.6]{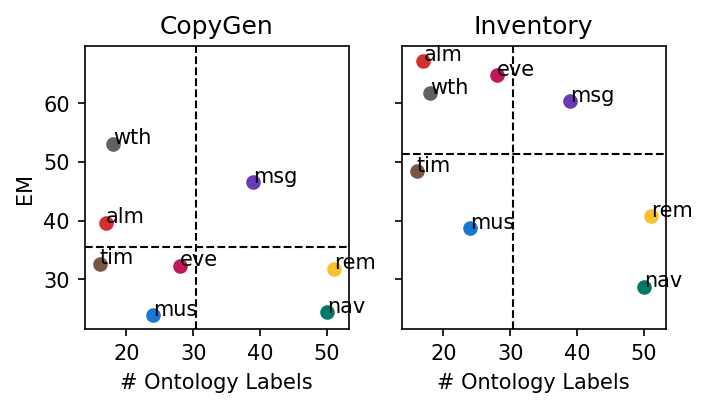}
    \caption{EM versus \% compositionality and \# ontology labels, two important characteristics of task-oriented domains at 1 SPIS. Horizontal lines show average EM, while vertical lines show average characteristics. Domains include alarm (alm), event (eve), messaging (msg), music (mus), navigation (nav), reminder (rem), timer (tim), and weather (wth).}
    \label{fig:domain_plot}
\end{figure}

\begin{table*}
\centering
\begin{tabular}{p{2.5cm}p{12cm}}
\toprule
Model & Utterance / Frame \\
\midrule
 & I need you to send a video message now \\
Index & \texttt{[IN:SEND\_MESSAGE ] } \\
\quad+ Type, Span & \texttt{[IN:SEND\_MESSAGE \hlblue{[SL:TYPE\_CONTENT \textrm{video} ]} ]}  \\
\midrule
 & Did I get any messages Tuesday on Twitter \\
Index & \texttt{[IN:GET\_MESSAGE \hlred{[SL:RECIPIENT \textrm{i} ] [SL:ORDINAL} \textrm{Tuesday} ] \hlred{[SL:TAG\_MESSAGE} \textrm{Twitter} ] ]} \\
\quad+ Type, Span & \texttt{[IN:GET\_MESSAGE \hlblue{[SL:DATE\_TIME} \textrm{Tuesday} ] \hlblue{[SL:RESOURCE} \textrm{Twitter} ] ]} \\
\midrule
 & Message Lacey and let her know I will be at the Boxer Rescue Fundraiser Saturday around 8 \\
Index & \texttt{[IN:SEND\_MESSAGE [SL:RECIPIENT \textrm{Lacey} ] [SL:CONTENT\_EXACT \textrm{I will be at the Boxer Rescue Fundraiser} \hlred{] [SL:GROUP} \textrm{Saturday around 8} ] ]} \\
\quad+ Type, Span & \texttt{[IN:SEND\_MESSAGE [SL:RECIPIENT \textrm{Lacey} ] [SL:CONTENT\_EXACT \textrm{I will be at the Boxer Rescue Fundraiser Saturday around 8} ] ]} \\
\bottomrule
\end{tabular}
\caption{\textbf{Comparing index only and index + type + span parsers.} In each row, we show the utterance and, for each model, its predicted frame; here, the index + type + span frames are always correct. For visualization of edit distance, we use \hlblue{} and \hlred{} to indicate additions and deletions, respectively.}
\label{tab:ablation-comparison}
\end{table*}

\begin{table}[t]
\centering
\setlength{\tabcolsep}{4pt}
\begin{tabular}{lrrrr}
\toprule
 & \multicolumn{4}{c}{SPIS} \\
\cmidrule(lr){2-5}
Component & 1 & 2 & 5 & 10 \\
\midrule
Index & 36.78 & 44.13 & 60.63 & 61.90 \\
\quad+ Type & 46.54 & 49.67 & 65.21 & 69.98 \\
\quad+ Span & \textbf{60.36} & \textbf{66.68} & \textbf{74.69} & \textbf{78.04} \\
\bottomrule
\end{tabular}
\caption{\textbf{Inventory ablation experiment results.} We benchmark the performance of three Inventory$_\textrm{LARGE}$ models on the messaging domain, each adding an intrinsic property to their inventories. Our full model, where each component consists of an index, type, and span, outperforms baselines by a wide margin.}
\label{tab:ablation}
\end{table}

Table~\ref{tab:topv2da-average-results} presents the EM of CopyGen and Inventory on TOPv2-DA averaged across 8 domains. We also present more fine-grained results in Table~\ref{tab:topv2da-domain-results}, breaking down EM by domain. From these tables, we draw the following conclusions:

\paragraph{Inventory consistently outperforms CopyGen in 1, 2, 5, and 10 SPIS settings.} On average, Inventory shows improvements across the board, improving upon CopyGen by at least +10 EM on each SPIS subset. Compared to CopyGen, Inventory is especially strong at 1 SPIS, demonstrating gains of +11 and +15 EM across the base and large variants, respectively. Furthermore, we see \textbf{Inventory$_\textrm{BASE}$ outperforms CopyGen$_\textrm{LARGE}$}, indicating our model's performance can be attributed to more than just the pre-trained weights and, as a result, carries more utility in compute-constrained environments. 

However, provided that these constraints are not a concern, Inventory makes better use of larger representations. Figure~\ref{fig:parser_gap} illustrates this by plotting the $\Delta$ EM between the base and large variants of both models. The delta is especially pronounced at 1 SPIS, where Inventory$_\textrm{BASE}$ $\rightarrow$ Inventory$_\textrm{LARGE}$ yields +12 EM but CopyGen$_\textrm{BASE}$ $\rightarrow$ CopyGen$_\textrm{LARGE}$ only yields +7 EM. Unlike CopyGen which requires fine-tuning extra parameters in a target domain, Inventory seamlessly integrates stronger representations without modification to the underlying architecture. This is an advantage: we expect our model to iteratively improve in quality with the advent of new pre-trained transformers.

\paragraph{Inventory also yields strong results when inspecting each domain separately.} TOPv2 domains typically have a wide range of characteristics, such as their compositionality or ontology size, so one factor we can investigate is how our model performs on a per-domain basis. Specifically, is our model generalizing across the board or overfitting to particular settings? Using the per-domain, large model, 1 SPIS results in Table~\ref{tab:topv2da-domain-results}, we analyze EM versus \textbf{\% compositionality} (fraction of nested frames) and \textbf{\# ontology labels} (count of intent and slots). Figure~\ref{fig:domain_plot} plots these relationships. A key trend we notice is Inventory improves EM in general, though better performance is skewed towards domains with 20\% compositionality and 20-30 ontology labels. This can be partially explained by the fact that domains with these characteristics are more empirically dominant in TOPv2, as shown by the proximity of the dots to the vertical bars. Domains like reminder and navigation are more challenging given the size of their ontology space, but Inventory still outperforms CopyGen by a reasonable margin.

\begin{table*}
\centering
\begin{tabular}{p{2.5cm}p{12cm}}
\toprule
Model & Utterance / Frame \\
\midrule
\multicolumn{2}{l}{Domain: Alarm} \\
\midrule
 & Delete my 6pm alarm \\
Inventory & \texttt{[IN:DELETE\_ALARM [SL:DATE\_TIME \textrm{6pm} ] ]} \\
Oracle & \texttt{[IN:DELETE\_ALARM \hlblue{[SL:ALARM\_NAME [IN:GET\_TIME} [SL:DATE\_TIME \textrm{6pm} ] \hlblue{] ]} ]}  \\
\midrule
\multicolumn{2}{l}{Domain: Event} \\
\midrule
 & Fun activities in Letchworth next summer \\
Inventory & \texttt{[IN:GET\_EVENT [SL:CATEGORY\_EVENT \textrm{\hlred{fun} activities} ] [SL:LOCATION \textrm{Letchworth} ] [SL:DATE\_TIME \textrm{next summer} ] ]} \\
Oracle & \texttt{[IN:GET\_EVENT [SL:CATEGORY\_EVENT \textrm{activities} ] [SL:LOCATION \textrm{Letchworth} ] [SL:DATE\_TIME \textrm{next summer} ] ]}  \\
\midrule
\multicolumn{2}{l}{Domain: Messaging} \\
\midrule
 & Message Candy to send me details for her baby shower \\
Inventory & \texttt{[IN:SEND\_MESSAGE \hlred{[SL:SENDER} \textrm{Candy} ] [SL:CONTENT\_EXACT \textrm{details for her baby shower} ] ]} \\
Oracle & \texttt{[IN:SEND\_MESSAGE \hlblue{[SL:RECIPIENT} \textrm{Candy} ] [SL:CONTENT\_EXACT \textrm{\hlblue{send me} details for her baby shower} ] ]}  \\
\midrule
\multicolumn{2}{l}{Domain: Navigation} \\
\midrule
 & What is the distance between Myanmar and Thailand \\
Inventory & \texttt{[IN:GET\_DISTANCE \hlred{[SL:UNIT\_DISTANCE} \textrm{Myanmar} ] \hlred{[SL:UNIT\_DISTANCE} \textrm{Thailand} ] ]} \\
Oracle & \texttt{[IN:GET\_DISTANCE \hlblue{[SL:SOURCE} \textrm{Myanmar} ] \hlblue{[SL:DESTINATION} \textrm{Thailand} ] ]}  \\
\midrule
\multicolumn{2}{l}{Domain: Reminder} \\
\midrule
 & Remind me that I have lunch plans with Derek in two days at 1pm \\
Inventory & \texttt{[IN:CREATE\_REMINDER [SL:PERSON\_REMINDED \textrm{me} ] [SL:TODO \textrm{I have lunch plans} ] [SL:ATTENDEE\_EVENT \textrm{Derek} ] [SL:DATE\_TIME \textrm{in two days} ] [SL:DATE\_TIME \textrm{at 1pm} ] ]} \\
Oracle & \texttt{[IN:CREATE\_REMINDER [SL:PERSON\_REMINDED \textrm{me} ] [SL:TODO \hlblue{[IN:GET\_TODO [SL:TODO} \textrm{lunch plans} ] [SL:ATTENDEE \textrm{Derek} ] \hlblue{] ]} [SL:DATE\_TIME \textrm{\hlblue{in two days} at 1pm} ] ]}  \\
\midrule
\multicolumn{2}{l}{Domain: Timer} \\
\midrule
 & Stop the timer \\
Inventory & \texttt{\hlred{[IN:DELETE\_TIMER} [SL:METHOD\_TIMER \textrm{timer} ] ]} \\
Oracle & \texttt{\hlblue{[IN:PAUSE\_TIMER} [SL:METHOD\_TIMER \textrm{timer} ] ]} \\
\midrule
\multicolumn{2}{l}{Domain: Weather} \\
\midrule
 & What is the pollen count for today in Florida \\
Inventory & \texttt{\hlred{[IN:GET\_WEATHER} [SL:WEATHER\_ATTRIBUTE \textrm{pollen} ] [SL:DATE\_TIME \textrm{for today} ] [SL:LOCATION \textrm{Florida} ] ]} \\
Oracle & \texttt{\hlblue{[IN:UNSUPPORTED\_WEATHER} [SL:WEATHER\_ATTRIBUTE \textrm{pollen \hlblue{count}} ] [SL:DATE\_TIME \textrm{for today} ] [SL:LOCATION \textrm{Florida} ] ]} \\
\bottomrule
\end{tabular}
\caption{\textbf{Error analysis of domain-specific inventory parsers.} In each row, we show the utterance and compare our inventory model's predicted frames to an oracle model's gold frames. For visualization of edit distance, we use \hlblue{} and \hlred{} to indicate additions and deletions, respectively.}
\label{tab:oracle-comparison}
\end{table*}

\subsection{Inventory Ablation}

Moving beyond benchmark performance, we now turn towards better understanding the driving factors behind our model's performance. From Section~\ref{sec:inventories}, recall each inventory component consists of an index, type, and span. The index is merely a unique identifier, while the type and span represent intrinsic properties of a label. Therefore, the goal of our ablation is to quantify the impact adding types and spans to inventories. Because conducting ablation experiments on each domain is cost-prohibitive, we use the messaging domain as a case study given its samples strike a balance between compositionality and ontology size.

We experiment with three Inventory$_\textrm{LARGE}$ models, where each model iteratively adds an element to its inventory components: (1) index only, (2) index and type, (3) index, type, and span. The results are shown in Table~\ref{tab:ablation}. Here, we see that while an index model performs poorly, adding types and spans improve performance across all subsets. At 1 SPIS, in particular, an index model improves by roughly +10 and +20 EM when types and spans are added, respectively. These results suggest that these intrinsic properties provide a useful inductive bias in the absence of copious training data.

In Table~\ref{tab:ablation-comparison}, we contrast the predictions of the index only (1) and index + type + span (3) models more closely, specifically looking at 1 SPIS cases where the frame goes from being incorrect to correct. We see a couple of cases where knowing about a label's intrinsic properties might help make the correct assessment during frame generation. The second example shows a scenario where our model labels ``tuesday'' as \texttt{SL:DATE\_TIME} rather than \texttt{SL:ORDINAL}. This distinction is more or less obvious when contrasting the phrases ``date time'' and ``ordinal'', where the latter typically maps to numbers. In the third example, a more tricky scenario, our model correctly labels the entire subordinate clause as an exact content slot. While partitioning this clause and assigning slots to its constituents may yield a plausible frame, in this instance, there is not much correspondence between \texttt{SL:GROUP} and ``saturday around 8''.

\section{Error Analysis}
\label{sec:error-analysis}

Thus far, we have demonstrated the efficacy of inventory parsers, but we have not yet conducted a thorough investigation of their errors. Though models may not achieve perfect EM in low-resource settings, they should ideally fail gracefully, making mistakes which roughly align with intuition. In this section, we assess this by combing through our model's cross-domain errors. Using Inventory$_\textrm{LARGE}$ models fine-tuned in each domain's 1 SPIS setting, we first manually inspect 100 randomly sampled errors to build an understanding of the error distribution. Then, for each domain, we select one representative error, and present the predicted and gold frame in Table~\ref{tab:oracle-comparison}.

In most cases, the edit distance between the predicted and gold frames is quite low, indicating the frames our models produce are fundamentally good and do not require substantial modification. We do not see evidence of erratic behavior caused by autoregressive modeling, such as syntactically invalid frames or extraneous subword tokens in the output sequence. Instead, most errors are relatively benign; we can potentially resolve them with rule-based transformations or data augmentation, though these are outside the scope of our work. Below, we comment on specific observations:

\paragraph{Frame slots are largely correct and respect linguistic properties.} One factor we investigate is if our model copies over utterance spans correctly, which correspond to arguments in an API call. These spans typically lie on well-defined constituent boundaries (e.g., prepositional phrases), so we inspect the degree to which this is respected. Encouragingly, the vast majority of spans our model copies over are correct, and the cases which are incorrect consist of adding or dropping modifiers. For example, in the event example, our model adds the adjective ``fun'', and in the weather example, our model drops the noun ``count''. These cases are relatively insignificant; they are typically a result of annotation inconsistency and do not carry much weight in practice. However, a more serious error we see is failing to copy over larger spans. For example, in the reminder example, $\texttt{SL:DATE\_TIME}$ corresponds to both ``in two days'' and ``at 1pm'', but our model only copies over the latter.

\paragraph{Predicting compositional structures is challenging in low-resource settings.} Our model also struggles with compositionality in low-resource settings. In both the alarm and reminder examples, our model does not correctly create nested structures, which reflect how slots ought to be handled during execution. Specifically, in the alarm example, because ``6pm'' is both a name and date/time, the gold frame suggests resolving the alarm in question before deleting it. Similarly, in the reminder example, we must first retrieve the ``lunch plans'' todo before including it as a component in the remainder of the frame. Both of these cases are tricky as they target prescriptive rather than descriptive behavior. Parsers often learn this type of compositionality in a data-driven fashion, but it remains an open question how to encourage this behavior given minimal supervision.

\paragraph{Ontology labels referring to ``concepts'' are also difficult.} Another trend we notice is our model predicts concept-based ontology labels with low precision. These labels require understanding a deeper concept which is not immediately clear from the surface description. A prominent example of this is the ontology label $\texttt{IN:UNSUPPORTED\_WEATHER}$ used to tag unsupported weather intents. To use this label, a parser must understand the distinction between in-domain and out-of-domain intents, which is difficult to ascertain from inventories alone. Other examples of this phenomenon manifest in the messaging and navigation domain with the slot pairs ($\texttt{SL:SENDER}$, $\texttt{SL:RECIPIENT}$) and ($\texttt{SL:SOURCE}$, $\texttt{SL:DESTINATION}$), respectively. While these slots are easier to comprehend given their intrinsic properties, a parser must leverage contextual signals and jointly reason over their spans to predict them.

\section{Related Work}
\label{sec:related-work}

Prior work improving the generalization of task-oriented semantic parsers can be categorized into two groups: (1) contextual model architectures and (2) fine-tuning and optimization. We compare and contrast our work along these two axes below.

\paragraph{Contextual Model Architectures.} \citet{bapna-2017-domain-scaling,lee-2018-zs-xlu,shah-2019-xdomain-slot} propose BiLSTMs which process both utterance and slot description embeddings, and optionally, entire examples, to generalize to unseen domains. Similar to our work, slot descriptions help contextualize what their respective labels align to. These descriptions can either be manually curated or automatically sourced. Our work has three key differences: (1) Inventories are more generalizable, specifying a format which encompasses multiple intrinsic properties of ontology labels, namely their types and spans. In contrast, prior work largely focuses on spans, and that too, only for slot labels. (2) Our model is interpretable: the decoder explicitly aligns inventory components and utterance spans during generation, which can aid debugging. However, slot description embeddings are used in an opaque manner; the mechanism through which BiLSTMs use them to tag slots is largely hidden. (3) We leverage a seq2seq framework which integrates inventories without modification to the underlying encoder and decoder. In contrast, prior work typically builds task-specific architectures consisting of a range of trainable components, which can complicate training.

\paragraph{Fine-tuning and Optimization.} Recently, low-resource semantic parsing has seen a methodological shift with the advent of pre-trained transformers. Instead of developing new architectures, as discussed above, one thrust of research tackles domain adaptation via robust optimization. These methods are typically divided between source and target domain fine-tuning. \citet{chen-2020-topv2} use Reptile \cite{nichol-2018-reptile}, a meta-learning algorithm which explicitly optimizes for generalization during source fine-tuning. Similarly, \citet{ghoshal-2020-loras} develop LORAS, a low-rank adaptive label smoothing algorithm which navigates structured output spaces, therefore improving target fine-tuning. Our work is largely orthogonal; we focus on redefining the inputs and outputs of a transformer-based parser, but do not subscribe to specific fine-tuning or optimization practices. Our experiments use MLE and Adam for simplicity, though future work can considering improving our source and target fine-tuning steps with better algorithms. However, one important caveat is both Reptile and LORAS rely on strong representations (i.e., BART$_\textrm{LARGE}$) for maximum efficiency, and typically show marginal returns with weaker representations (i.e., BART$_\textrm{BASE}$). In contrast, even when using standard practices, both the base and large variants of our model perform well, indicating our approach is more broadly applicable.

\section{Conclusion}
\label{sec:conclusion}

In this work, we present a seq2seq-based, task-oriented semantic parser based on inventories, tabular structures which capture the intrinsic properties of an ontology space, such as label types (e.g., ``slot'') and spans (e.g., ``time zone''). Our approach is both simple and flexible: we leverage out-of-the-box, pre-trained transformers with no modification to the underlying architecture. We chiefly perform evaluations on TOPv2-DA, a benchmark consisting of 32 low-resource experiments across 8 domains. Experiments show our inventory parser outperforms classical copy-generate parsers by a wide margin and ablations illustrate the importance of types and spans. Finally, we conclude with an error analysis, providing insight on the types of errors practitioners can expect when using our model in low-resource settings.

\bibliography{custom}
\bibliographystyle{acl_natbib}

\end{document}